\documentclass{amsproc}
\usepackage{amsaddr}
\usepackage[utf8]{inputenc}
\usepackage{booktabs}
\usepackage{tikz}
\usepackage{geometry}
\usepackage{setspace}

\title{Using Language Models to Detect Alarming Student Responses}
\author{Chirstopher Ormerod, Milan Patel, and Harry Wang}

\address{Cambium Assessment \\
1000 Thomas Jefferson St NW, Washington\\
District of Columbia, 20007, United States}

\email{christopher.ormerod@cambiumassessment.com}


\begin{document}

\maketitle

\begin{abstract}
This article details the advances made to a system that uses artificial intelligence to identify alarming student responses. This system is built into our assessment platform to assess whether a student's response indicates they are a threat to themselves or others. Such responses may include details concerning threats of violence, severe depression, suicide risks, and descriptions of abuse. Driven by advances in natural language processing, the latest model is a fine-tuned language model trained on a large corpus consisting of student responses and supplementary texts. We demonstrate that the use of a language model delivers a substantial improvement in accuracy over the previous iterations of this system.  
\end{abstract}

\section{Introduction}

Automated Text Scoring (ATS) refers to using artificial intelligence (AI) to approximate the assessment of constructed text responses. Despite its potential for reducing costs, ensuring consistent scores, and minimizing bias, there are still very real and valid concerns about the complete removal of human oversight from the scoring process. In particular, this article concerns instances where the constructed response suggests that the student poses a risk to themselves or others. These are situations where it is necessary for a person to intervene to ensure the safety of everyone in the school community. We refer to responses of this nature as an Alarming Student Response (ASR), which may include threats of violence, severe depression, suicide risks, and descriptions of abuse \cite{burkhardt_rubric_2021}. This program is incredibly important, especially given the regularity and severity of school shootings.

While some testing agencies ensure that all responses are reviewed by a person to detect ASRs, screening for these types of responses can be very time-consuming, especially when millions of responses are received daily. Furthermore, the concerning situations associated with these responses can be time-sensitive. This study concerns the advancements made to our human-AI hybrid system in which the same AI used in ATS systems are used to prioritize a small collection of responses for human review. By integrating AI into the detection of ASRs, the time required to provide an appropriate response can be significantly reduced. The program in place, and the infrastructure and protocols around this piece of AI, aim to provide the fastest possible response to life-threatening situations. 

We typically categorize ATS into two main classes; Automated Essay Scoring (AES) and Automated Short Answer Scoring (ASAS). The development of AES can be traced back to the 1960s when researchers started exploring the use of computers to analyze and evaluate natural language. The first AES system, Project Essay Grade (PEG), was developed in 1966 by Ellis Page and used a set of rules to analyze the structure and content of an essay \cite{page_project_2003}. The PEG system is based on a concept known as Bag-of-Words (BoW), which uses the frequencies of keywords in addition to hand-crafted features. A BoW model was the first instance of ATS methods used to detect ASRs.

It is well known that frequency-based approaches, such as BoW, can be very brittle when handling ASAS. This is why reliable systems for ASAS were developed much later \cite{leacock_c-rater_2003}. Over the years, researchers have refined the algorithms and techniques used in AES and ASAS. With the increasing sophistication of natural language processing (NLP) researchers began to use machine learning techniques such as neural networks \cite{taghipour_neural_2016, dong_attention-based_2017}. In 2018, a systematic study was conducted on the effectiveness of recurrent neural network architectures in the detection of ASRs \cite{ormerod_neural_2018}. The study considered two main recurrent neural network (RNN) architectures; Long-Short-Term Memory (LSTM) networks and Gated Recurrent Units \cite{cho_learning_2014} with and without attention \cite{bahdanau_neural_2016}. As a result of \cite{ormerod_neural_2018}, the BoW model used to detect ASRs was replaced with a two-layer bidirectional LSTM with attention.

More recently, the field has embraced the use of language models in both AES \cite{ormerod_automated_2021, uto_automated_2020} and ASAS \cite{ormerod_short-answer_2022, ormerod_automated_2022}. It stands to reason that the detection of ASRs could greatly benefit from language models. This article demonstrates significant improvements from using language models to detect ASRs over previous methods.

This article is organized into the following sections: we present the data and give a brief overview of how the model works, and how we train it to detect ASRs in \S \ref{sec:method}. In \S \ref{sec:results} we detail the improvements over two baselines used in previous generations of the program.

\section{Method}\label{sec:method}

To understand the context of the model, we give a brief overview of how the model fits into the broader system used to detect ASRs. We then consider the data used to train and validate the model, we then consider the model itself, why we chose it, and how it was trained.

\subsection{The system}

Given the nature of the problem, it is not reasonable to expect that a machine-learning model is solely responsible for the detection of alarming student responses. The model that detects alarming student responses is part of a larger program that is built into an online assessment program. When a student submits a response, our system divides the response into multiple fragments, each of which is processed by the model, which classifies the response as either a normal student response or a response that should be routed to another system for human review. During the 2018-2019 school year, this system processed almost 82 million fragments, which, during the peak testing season can reach up to 7 million responses a day. This means that efficiency is a key concern. The faster the system can process responses, the faster the human review can be completed, and ultimately, the faster human intervention can take place. 

\begin{table}[!ht]
    \centering
    \begin{tabular}{p{5cm} p{5cm} p{4cm}} \toprule
        Category & Details & Examples \\ \midrule
         Harm to self & Suicidal thoughts or actions &  I wanna kill myself\\
                      & Self harming thoughts or actions & I cut a lot\\
                      & Eating disorder & \\
                      & Drug Use & \\
                      
         Harm to another & Threat or admission of violence & I hit my girlfriend\\
          & Threat of sexual assault & All (PC) must die \\
          & Threatening hate speach & I want a sniper rifle\\
         Harm from another & Report of abuse & My dad beats me \\
         & Report of sexual assault & I get bullied \\
         & Bullying & \\
         Severe depression and/or trauma & Ongoing or unresolved & Please kill me \\ 
         &  & I want to die \\
         && I wish I was dead\\
         Specific serious request for help & Not test related & I hate my life, please help \\
         && help me or kill me \\ \bottomrule
    \end{tabular}
    \caption{A rubric for the detection of ASRs as seen in \cite{burkhardt_rubric_2021}. PC stands for Protected Class.}
    \label{tab:rubric}
\end{table}

A trained team of reviewers is responsible for reading the response and determining whether the response is a true ASR or simply a false positive. A full rubric for how they are assessed was featured in \cite{burkhardt_rubric_2021}. A summary of this rubric is presented in Table \ref{tab:rubric}. Once a response is deemed to be an ASR, the response and any identifying information are sent via a secure platform to the appropriate authorities. 

There are some responses that may be genuine ASRs that are not correctly classified by the model, however, depending on where and how the response was entered, those responses are also scored by hand-scorers and are subjected to a set of protocols in case they are alarming in nature. 

\subsection{Data}

For any modeling to be effective, it is crucial that the data is carefully considered and appropriate for the problem at hand. We use an updated version of the corpus that was used in \cite{ormerod_neural_2018}. The system in place assesses texts from various sources, such as short answers, essay responses, and comments made by students. We estimate that true ASRs are approximately 0.012\% of all responses \cite{ormerod_neural_2018}, which creates an imbalanced and difficult classification problem. To address this issue, we supplement actual ASRs with responses from a diverse set of open online forums and other texts that are similar to student responses. By including this supplementary data, our dataset heavily oversamples the ASRs, resulting in approximately 2.08\% of all responses being ASRs in our training set, which is still about 100 times the frequency that they appear in production, but makes the classification problem more manageable. An outline of this data is detailed in Table \ref{traindat}.

In this project, we do not adhere to standard machine learning practices for segmenting our data into different splits because our objective is not to enhance accuracy. Rather, in the operational setting, the engine must classify a conservatively defined fixed percentage of the overall population for review. In practice, our aim is to increase the number of true positives within that fixed percentage even if it means that we have a very large number of false positives. As a result, the model validation process is very different. For a given response, we use the model's output probability as a measure of the severity of the response and establish thresholds that align with the fixed percentages.

\begin{table}[!ht]
\begin{tabular}{l  l | r r r} \toprule
Category & & ASR & Normal  & Total \\ \toprule
Training & Student & 20,409 & 1,214,381  &  1,234,790 \\
 & Supplementary & 5,476 & 4,122 &  9,598  \\ \midrule 
 & Total & 25,885 & 1,218,503&  1,244,388\\\bottomrule
\end{tabular}
\caption{This data represents the training data used in this study.\label{traindat}}
\end{table}

The model was trained using supervised learning using the labeled data. This corpus is segmented into a training set which is 80\% of the data, and a random development set, which is 20\% of the data. In addition to the training data, we curated a collection of exactly 1 million unlabeled texts which we call the threshold data. The threshold data is a collection of data randomly chosen and representative of typical responses. The motivation for using the threshold data is that we obtain the distribution of model outputs found in live operations. We can set a cutoff point corresponding to a fixed percentage of all responses we wish to send for review.

Our validation sample consists of a set of one thousand ASRs that have been reviewed by human experts. Using the threshold data, an associated cutoff value, and the percentage of responses that we can review, we aim to determine the number of ASRs that will be correctly classified by the model as being for review. The percentage of correctly classified ASRs from the validation sample is our measure of the effectiveness of the model for that percentage. Typically we want to know the effectiveness of the model for a number of viable percentage values, which informs the decision regarding what percentage of all responses we should review.

\subsection{Previous Benchmarks}

The Bag-of-Words (BoW) model is based on the term-frequency inverse-document-frequency ($\mathrm{tf}$-$\mathrm{idf}$) matrix. Suppose we have a set of training data, $D$, which contains words from a vocabulary, $V$. We can summarize the word-frequency information in $D$ in a matrix, $X = (x_{v,d})$, where $d \in D$ and $v \in V$. We define the term frequency matrix 
\begin{equation}
    \mathrm{tf}(v,d) = x_{v,d}/ \left( \sum_{u} x_{u,d} \right)
\end{equation}
which encodes the proportions of each word in the document. The other component is the inverse document frequency term, given by 
\begin{equation}
\mathrm{idf}(v) = \mathrm{log} \dfrac{|D|}{|\{d \in D : v \in D\}|}.
\end{equation}
Together, the $\mathrm{tf}$-$\mathrm{idf}$ matrix is given by
\begin{equation}
T_{v,d} = \mathrm{tf}(v,d)\cdot \mathrm{idf}(v)
\end{equation}
This matrix still has far too many rows for a meaningful analysis, so what is often done is we perform a latent semantic analysis (LSA), where a fixed number of dominant eigenvectors of the  $\mathrm{tf}$-$\mathrm{idf}$ matrix summarizes the key features of the space. By discarding all but the fixed number of dominant eigenvectors, we obtain a transformation from the set of documents to some fixed dimensional vector space. The classification pipeline that uses this vector space as the input into a traditional logistic regression is called a Bag-of-Words classifier. What this approach operates on is a collection of important words in the student response, which does not capture the true semantics of the response. We trained a BoW model using 500 eigenvectors on the training data for this study.

The other classifier, a recurrent neural network, is based on the Long-Short-Term-Memory unit \cite{hochreiter_long_1997}. Firstly, all words are mapped to a vector space via an embedding, such as the GloVe embedding \cite{pennington_glove_2014}, and then are sequentially used as input into the recurrent neural network. Within the neural network, then input of a recurrent unit is a word vector and the memory state from the previous iteration where the initial memory state is zero. The memory state within a recurrent neural network allows for information to persist and be used in any final classification. We use the final output of the LSTM as input into a linear classifier, whose outputs are interpreted as log probabilities for a classification. 

\begin{figure}[!ht]
\begin{tikzpicture}[scale=1.3]
\node [draw, fill=blue!10,circle,minimum size=.9cm] (a) at (-3,1) {${}_a$};
\node [draw, fill=blue!10,circle,minimum size=.9cm] (0) at (0,1) {$0$};
\node [draw, fill=blue!10,circle,minimum size=.9cm] (a1) at (1,1) {${}_{a_1}$};
\node [draw, fill=blue!10,circle,minimum size=.9cm] (a2) at (2,1) {${}_{a_2}$};
\node [draw, fill=blue!10,circle,minimum size=.9cm] (anm1) at (4,1) {${}_{a_{n-1}}$};
\node [draw, fill=blue!10,circle,minimum size=.9cm] (an) at (5,1) {${}_{a_n}$};
\node at (3,1) {$\ldots$};
\node at (3,0) {$\ldots$};
\node at (3,2) {$\ldots$};
\node [draw, fill=red!10,circle,minimum size=.9cm] (x) at (-3,0) {${}_x$};
\node [draw, fill=red!10,circle,minimum size=.9cm] (x1) at (1,0) {${}_{x_1}$};
\node [draw, fill=red!10,circle,minimum size=.9cm] (x2)at (2,0) {${}_{x_2}$};
\node [draw, fill=red!10,circle,minimum size=.9cm] (xnm1) at (4,0) {${}_{x_{n-1}}$};
\node [draw, fill=red!10,circle,minimum size=.9cm] (xn) at (5,0) {${}_{x_n}$};

\node [draw, fill=green!10,circle,minimum size=.9cm] (h) at (-3,2) {${}_{h_n}$};
\node (h1) at (1,2) {${}_{h_1}$};
\node (h2)at (2,2) {${}_{h_2}$};
\node (hnm1) at (4,2) {${}_{h_{n-1}}$};
\node [draw, fill=green!10,circle,minimum size=.9cm] (hn) at (5,2) {${}_{h_n}$};

\draw[->,thick] (x)--(a);
\draw[->,thick] (x1)--(a1);
\draw[->,thick] (x2)--(a2);
\draw[->,thick] (xnm1)--(anm1);
\draw[->,thick] (xn)--(an);
\draw[->,thick] (0)--(a1);
\draw[->,thick] (a1)--(a2);
\draw[->,thick] (a2) -- (2.5,1);
\draw[->,thick] (3.5,1) -- (anm1);
\draw[->,thick] (anm1) -- (an);
\draw[->,thick] (a1) -- (h1);
\draw[->,thick] (a2) -- (h2);
\draw[->,thick] (anm1) -- (hnm1);
\draw[->,thick] (an) -- (hn);
\draw[->,thick] (a) -- (h);
\end{tikzpicture}
\caption{\label{RNN}When we unfold an RNN, we express it as a sequence of cells each accepting, as input, an element of the sequence. The output of the RNN is the output of the last state.}
\end{figure}
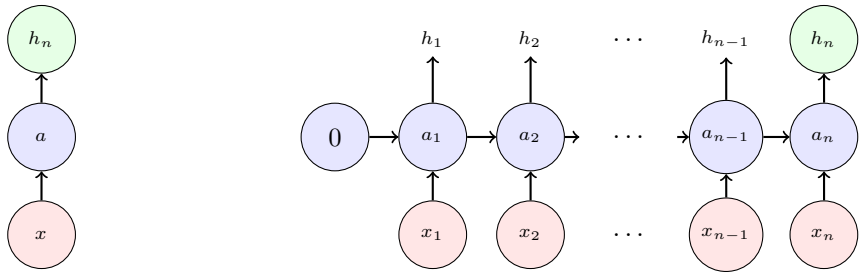

The first modification of the basic RNN architecture is to use the states of the RNN as input into another layer of recurrent units. This practice is known as stacking RNNs \cite{hochreiter_long_1997}. A second modification is to segment the recurrent units into two groups; one in which we use the sequence of inputs and the other in which we use the reversed sequence. For obvious reasons, the resulting structure is called a bidirectional RNN. Lastly, we can subject the outputs of the RNN to an attention mechanism \cite{bahdanau_neural_2016}, which is a way we can appropriately focus on important aspects of the input to classification and disregard those aspects that are not as important. All these modifications to the basic RNN structure are beneficial to many downstream tasks. An attempt to quantify the benefit of these modifications to the detection of ASRs appears in \cite{ormerod_neural_2018} in addition to the performance of the final RNN model we use to compare the language model. We used the training sample to train a bidirectional two-layer LSTM model with 512 hidden units (in each direction) with attention for this study.

\subsection{Modeling Details}

The introduction of transformer-based language models has sparked a revolution in natural language processing. Among the first of these models was the GPT model \cite{radford_improving_2018} and the BERT model \cite{devlin_bert_2019}. These models made waves by establishing new state-of-the-art benchmarks on a standard set of tasks designed to push the limits of natural language processing models \cite{wang_glue_2019}. The underlying premise is that one can improve downstream tasks on supervised data, where the corpora are limited in size, by pretraining the model on unsupervised data, where the corpora can be as large as one needs. One uses pretraining to endow the model with a basic understanding of language based on a large dataset of text. Examples of datasets used are the Book Corpus \cite{zhu_aligning_2015}, which is about 4.5GB, Wikipedia dumps, which has 40GB in English alone, and the C4 dataset derived from the common crawl, which is about 750GB. Training such models is a huge investment of time, money, and computing power, which ultimately equates to a sizable carbon footprint \cite{luccioni_estimating_2022}. 

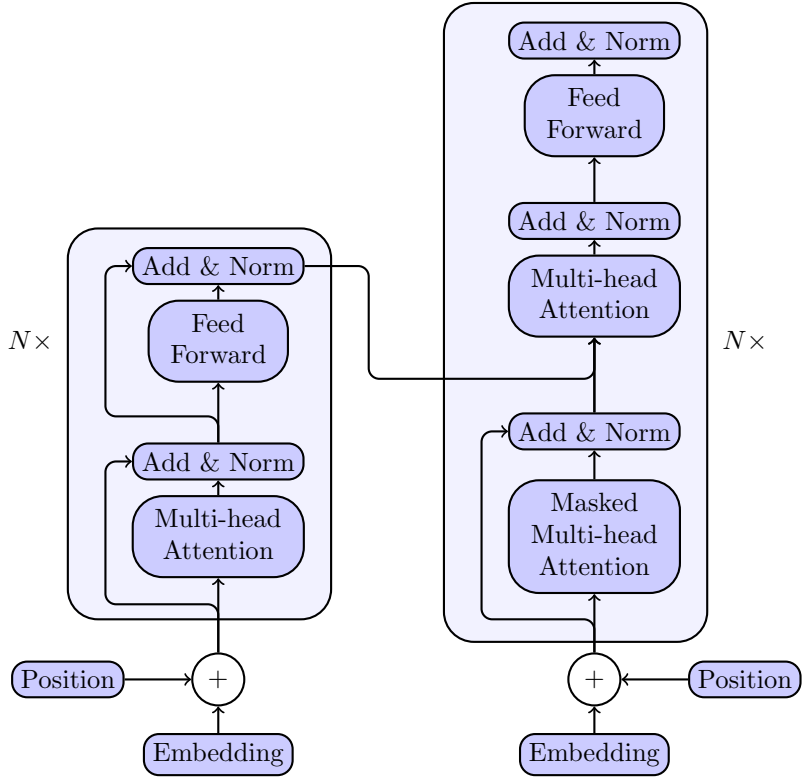
\begin{figure*}[!ht]
    \centering
    \begin{tikzpicture}[scale=1.0]
    \node[draw = black, thick, rectangle, rounded corners= 2mm, fill = blue!20](inp1) at (2,-1.5) {Embedding};
    \node[draw = black, thick, rectangle, rounded corners= 2mm, fill = blue!20](pos1) at (0,-.5) {Position};
    \node[circle, thick, draw=black](add1) at (2,-0.5) {$+$};
    \draw[thick,rounded corners= 4mm, fill = blue!5] (0,0.3) rectangle (3.5,5.5);
    \node[draw = black, thick, rectangle, rounded corners= 4mm, fill = blue!20](mh1) at (2,1.4) {\begin{tabular}{c}Multi-head\\ Attention\end{tabular}};
    \node[draw = black, thick, rectangle, rounded corners= 2mm, fill = blue!20](an1) at (2,2.4) {Add \& Norm};
    \node[draw = black, thick, rectangle, rounded corners= 4mm, fill = blue!20](ff1) at (2,4) {\begin{tabular}{c}Feed\\ Forward\end{tabular}};
    \node[draw = black, thick, rectangle, rounded corners= 2mm, fill = blue!20](an2) at (2,5) {Add \& Norm};
    \node[draw = black, thick, rectangle, rounded corners= 2mm, fill = blue!20](inp2) at (7,-1.5) {Embedding};
    \node[draw = black, thick, rectangle, rounded corners= 2mm, fill = blue!20](pos2) at (9,-.5) {Position};
    \node[circle, thick, draw=black](add2) at (7,-0.5) {$+$};
    \draw[thick,rounded corners= 4mm, fill = blue!5] (5,0) rectangle (8.5,8.5);
    \node[draw = black, thick, rectangle, rounded corners= 4mm, fill = blue!20](mmh1) at (7,1.4) {\begin{tabular}{c}Masked \\ Multi-head\\ Attention\end{tabular}};
    \node[draw = black, thick, rectangle, rounded corners= 2mm, fill = blue!20](an3) at (7,2.8) {Add \& Norm};
    \node[draw = black, thick, rectangle, rounded corners= 4mm, fill = blue!20](mh2) at (7,4.6) {\begin{tabular}{c}Multi-head\\ Attention\end{tabular}};
    \node[draw = black, thick, rectangle, rounded corners= 2mm, fill = blue!20](an4) at (7,5.6) {Add \& Norm};
    \node[draw = black, thick, rectangle, rounded corners= 4mm, fill = blue!20](ff2) at (7,7) {\begin{tabular}{c}Feed\\ Forward\end{tabular}};
    \node[draw = black, thick, rectangle, rounded corners= 2mm, fill = blue!20](an5) at (7,8) {Add \& Norm};
    \draw[thick,->] (pos1) -- (add1);
    \draw[thick,->] (inp1) -- (add1);
    \draw[thick,->] (add1) -- (mh1);
    
    \draw[thick,->] (pos2) -- (add2);
    \draw[thick,->] (inp2) -- (add2);
    \draw[thick,->] (add2) -- (mmh1);

    \draw[thick,->] (mh1) -- (an1);
    \draw[thick,->] (an1) -- (ff1);
    \draw[thick,->] (ff1) -- (an2);
    \draw[thick,->, rounded corners=4pt] (add1) -- (2,.5)--(0.5,.5) |- (an1);
    \draw[thick,->, rounded corners=4pt] (an1) -- (2,3) -- (.5,3) |- (an2);
    \draw[thick,->, rounded corners=4pt] (an2) -- (4,5) -- (4,3.5) -| (mh2);
    \draw[thick,->, rounded corners=4pt] (add2) -- (7,.3) -- (5.5,.3) |- (an3);
    \draw[thick,->, rounded corners=4pt] (mmh1) -- (an3);
    \draw[thick,->, rounded corners=4pt] (an3) -- (mh2);
    \draw[thick,->, rounded corners=4pt] (mh2) -- (an4);
    \draw[thick,->, rounded corners=4pt] (an4) -- (ff2);
    \draw[thick,->, rounded corners=4pt] (ff2) -- (an5);
    \draw[thick,->, rounded corners=4pt] (an3) -- (mh2);
    \node at (9,4) {$N\times $};
    \node at (-.5,4) {$N\times $};
    \end{tikzpicture}
    \caption{This is the basic architecture of a transformer-based model \cite{vaswani_attention_2017}. The left block of transformers is the encoder while the right block of $N$ layers is the decoder.}
    \label{fig:Transformer}
\end{figure*}

Models are typically pretrained on these huge corpora to be one of two types of models: generative models that are trained to perform next-word prediction, like GPT \cite{radford_improving_2018}, or masked models that are trained to predict masked words, like BERT \cite{devlin_bert_2019}. There are some variations on this, such as the adding sentence ordering to the loss function, or adversarial training mechanisms like Electra \cite{clark_electra_2020} or DeBERTa \cite{he_debertav3_2021}, however, the vast majority of language models available in standard libraries are masked-word models \footnote{See https://huggingface.co/docs/transformers.}. These models are all variations of the same transformer architecture described in \cite{vaswani_attention_2017}. This architecture is presented in Figure \ref{fig:Transformer}.

At the heart of the transformer-based model is the idea of an attention mechanism. From a mathematical perspective, we define a query matrix, $Q$, a key matrix, $K$ and a value matrix, $V$, which are all normalized linear transformations of the input matrix. Attention is defined as
\begin{equation}\label{eq:attn}
Attention(Q,K,V) = \mathrm{softmax}\left( \dfrac{QK^{T}}{\sqrt{d}}\right) V,
\end{equation}
where $d$ is the dimension of the query and key values. The softmax component of \eqref{eq:attn} is referred to as the attention matrix. The multi-headed attention that features in Figure \ref{fig:Transformer} operates on a decomposition of the input space into a disjoint union of subspaces and computes attention as it applies to each subspace \cite{devlin_bert_2019}.

Intuitively, attention and multi-headed attention are mechanisms that enable a model to selectively focus on certain parts of the input data while disregarding others. This mechanism is inspired by how humans selectively attend to relevant information and ignore irrelevant information. In neural networks, attention works by assigning weights to different parts of the input data, which indicates how important each part is to the training task. These weights are learned during the training process. During inference, the attention weights are used to compute a weighted sum of the input features, where the weights reflect the relevance of each feature for the task. This enables the model to selectively attend to the most informative features, which can improve its performance on the task. Self-attention is a key component of the Transformer architecture, which allows the model to capture long-range dependencies and contextual information within a sequence of input tokens, such as words in a sentence. In a transformer-based language model, multiple layers of self-attention are used on the entire sequence of input tokens at once, rather than attending to a fixed-size window of tokens as in traditional recurrent neural networks. 

The model we use for this task is a small and very efficient language model based on the Electra architecture \cite{clark_electra_2020}. The model follows a very similar architecture to the base BERT model where the hidden size, the feed-forward layer, and the number of attention heads are a third of those found in the base BERT model. Secondly, the small Electra model has a much smaller embedding facilitated by a linear layer between the embedding and the transformer layers. The more substantial change is the adversarial training scheme. The model is part of a pair of models, one trained to generate masked tokens, like BERT, and the other trained to distinguish between generated and original tokens. 

In classification tasks, such as the Corpus of Linguistic Acceptability (CoLA) \cite{warstadt_neural_2019} and the Stanford Sentiment Treebank (SST) \cite{socher_recursive_nodate}, the small Electra model delivers strong performance despite its small size. In fact, the small Electra model has a better performance on the Kaggle Automated Student Assessment Prize essay dataset \cite{ormerod_automated_2021, shermis_state---art_2014} and the Short Answer dataset \cite{ormerod_short-answer_2022, shermis_contrasting_2015}. Among the collection of all pretrained language models available using standard libraries \cite{wolf_huggingfaces_2020}, very few models are as computationally efficient while still delivering strong performance in classification tasks. A comparison with some of the relevant models to this study is presented in Table \ref{tab:glue_benchmarks}.

\begin{table}[]
    \centering
    \begin{tabular}{l| c c c c c c c c c c} \toprule
     & Params &  CoLA & SST & MRPC & STS & QQP & MNLI & QNLI & RTE & Avg\\ \midrule
BiLSTM + Attn & & 15.7 & 85.8 & 68.5 & 59.3 & 83.5 & 74.2 & 77.2 & 51.9 & 63.9\\
GPT  & 117M & 45.4 & 91.3 & 75.7 & 80.0 & 88.5 & 82.1 & 88.1 & 56.0 & 75.9\\
BERT (base) & 110M & 52.1 & 93.5 & 84.8 & 85.8 & 89.2 & 84.6 & 90.5 &66.4& 80.9\\ \
Electra (small) & 13M & 54.6 & 89.1 & 83.7 & 80.3 & 88.0 & 79.7 & 87.7 & 60.8 & 78.0\\ 
\bottomrule

    \end{tabular}
    \caption{Baseline performance on the GLUE task test sets. For more information regarding these tests, we refer to \cite{wang_glue_2019}.}
    \label{tab:glue_benchmarks}
\end{table}

After choosing a model, the next step is to apply it. Electra has input limits, so the number of operations required to compute attention grows quadratically with length. As a result, most models are limited to 512 tokens. We can train the engine on segments of 256 sub-word tokens to get around this, where ASRs are given a label of 1 and normal text is given a label of 0. In this way, we can interpret the model's output for a fragment as a measure of the level of concern of the response.

Specific references to self-harm or threats made to other teachers do not necessarily require long-term dependencies in the same way that an essay might. This is one of the key observations regarding ASRs. We can use this to our advantage by training the engine on segments of 256 sub-word tokens. This is because what typically makes ASRs alarming is isolated to a few sentences.

We fine-tuned the pretrained Electra model on our labeled text using the Adam optimizer with a weight decay mechanism \cite{loshchilov_decoupled_2019}, a learning rate of $2.5 \times 10^{-5}$, a batch size of 32, and a linear learning rate scheduler. Given the size of the data, we trained for 2 epochs on a T4 graphics card. At inference, we divide any fragment into segments of length 256 with an overlap of 32 sub-word tokens. The final score is the maximum model output over all the segments. We combine this with the Open Neural Network Exchange (ONNX) \footnote{https://github.com/onnx/onnx}, which is an optimized execution platform. With the advancements in GPU technology, and the availability of standard libraries supported by ONNX, our system is able to classify all responses even with heavy loads within 1-4 hours, whereas larger systems like BERT might take approximately 3 times longer on equivalent hardware with comparable efficacy.

\section{Results}\label{sec:results}

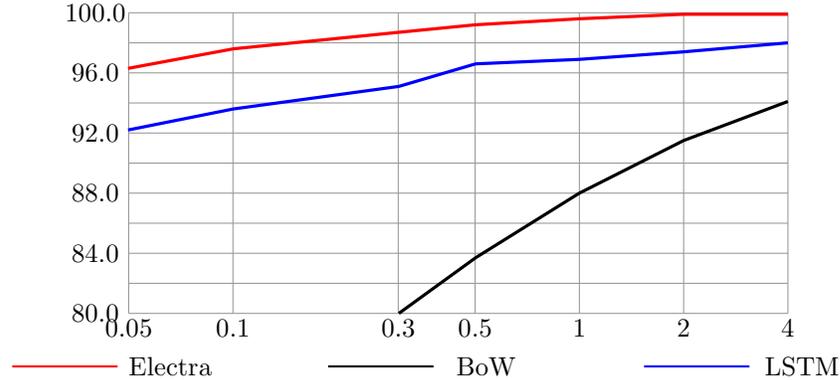
\begin{figure}[!ht]
\begin{center}
\begin{tikzpicture}[xscale=2, yscale=20]
\draw[black!40] (-2.995732273553991,.8) -- (-2.995732273553991,1);
\node at (-2.995732273553991,.79) {0.05};
\draw[black!40] (-2.3025850929940455,.8) -- (-2.3025850929940455,1);
\node at (-2.3025850929940455,.79) {0.1};
\draw[black!40] (-1.2039728043259361,.8) -- (-1.2039728043259361,1);
\node at (-1.2039728043259361,.79) {0.3};
\draw[black!40] (-0.6931471805599453,.8) -- (-0.6931471805599453,1);
\node at (-0.6931471805599453,.79) {0.5};
\draw[black!40] (0.0,.8) -- (0.0,1);
\node at (0.0,.79) {1};
\draw[black!40] (0.6931471805599453,.8) -- (0.6931471805599453,1);
\node at (0.6931471805599453,.79) {2};
\draw[black!40] (1.3862943611198906,.8) -- (1.3862943611198906,1);
\node at (1.3862943611198906,.79) {4};
\draw[black!40] (-2.995732273553991,0.8) -- (1.3862943611198906,0.8);
\node at (-3.2188758248682006,0.8) {80.0};
\draw[black!40] (-2.995732273553991,0.8200000000000001) -- (1.3862943611198906,0.8200000000000001);
\draw[black!40] (-2.995732273553991,0.8400000000000001) -- (1.3862943611198906,0.8400000000000001);
\node at (-3.2188758248682006,0.8400000000000001) {84.0};
\draw[black!40] (-2.995732273553991,0.8600000000000001) -- (1.3862943611198906,0.8600000000000001);
\draw[black!40] (-2.995732273553991,0.8800000000000001) -- (1.3862943611198906,0.8800000000000001);
\node at (-3.2188758248682006,0.8800000000000001) {88.0};
\draw[black!40] (-2.995732273553991,0.9000000000000001) -- (1.3862943611198906,0.9000000000000001);
\draw[black!40] (-2.995732273553991,0.9200000000000002) -- (1.3862943611198906,0.9200000000000002);
\node at (-3.2188758248682006,0.9200000000000002) {92.0};
\draw[black!40] (-2.995732273553991,0.9400000000000002) -- (1.3862943611198906,0.9400000000000002);
\draw[black!40] (-2.995732273553991,0.9600000000000002) -- (1.3862943611198906,0.9600000000000002);
\node at (-3.2188758248682006,0.9600000000000002) {96.0};
\draw[black!40] (-2.995732273553991,0.9800000000000002) -- (1.3862943611198906,0.9800000000000002);
\draw[black!40] (-2.995732273553991,1.0) -- (1.3862943611198906,1.0);
\node at (-3.2188758248682006,1.0) {100.0};
\draw[very thick] (-1.2,.80)--(-.69,.837) -- (0.0,.880) -- (.693,.915) -- (1.386,.941);
\draw[very thick, red](-3,0.963) -- (-2.3,.976) -- (-1.2,.987)--(-.69,.992) -- (0.0,.996) -- (.693,.999) -- (1.386,.999);
\draw[very thick, blue](-3,0.922) -- (-2.3,.936) -- (-1.2,.951)--(-.69,.966) -- (0.0,.969) -- (.693,.974) -- (1.386,.980);
\end{tikzpicture}

\begin{tikzpicture}[scale=.7]
\draw[red,thick] (0,0) -- (2,0);
\node at (3,0) {Electra};
\draw[black,thick] (6,0) -- (8,0);
\node at (9,0) {BoW};
\draw[blue,thick] (12,0) -- (14,0);
\node at (15,0) {LSTM};
\end{tikzpicture}
\caption{A graph of the approximate percentage of ASRs caught against the logarithm of the approximate percent of responses flagged.\label{fig:graph}}
\end{center}
\end{figure}

To evaluate the results, recall that our measure of efficacy is the percentage of ASRs in the validation sample that would be sent for review if some fixed percentage of all responses are sent for review. If we denote the efficacy by $E$, then $E$ is a function of the fixed percentage, which we denote $p$. We evaluate $E$ at values that make sense from an operational standpoint, which range from 0.05 to 4 percent. A table of efficacy values for our latest Electra model in addition to the bidirectional LSTM with attention and a BoW model has been presented in Table \ref{tab:efficacy_values}. Figure \ref{fig:graph} is a graphical representation of the efficacy as a function of the percentage, $p$ (on a logarithmic scale).

\begin{table}[]
    \centering
    \begin{tabular}{l|c c c c c c c c} \toprule
        Model  &  0.05 & 0.1 & 0.3 & 0.5 & 1 & 2 & 4\\ \midrule
        BoW + LSA + Logistic Regression & 61.3 & 69.8 & 80.0 & 83.7 & 88.0 &  91.5 & 94.1\\
    Bidirectional LSTM + Attention & 92.2 & 93.6 & 95.1 & 96.6 & 96.9 & 97.4 & 98.0 \\ 
    Electra small & 96.3 & 97.6 & 98.7 & 99.2 & 99.6 & 99.9 & 99.9 \\ \bottomrule
    \end{tabular}
    \caption{The efficacy values of the three different iterations of the systems trained for this study.}
    \label{tab:efficacy_values}
\end{table}

We see that the efficacy of the Electra model is on par at $p=0.1$ with the efficacy of the RNN at $2\%$. This would signify a twenty-fold reduction in the number of fragments that are required for review to catch the same number of ASRs.

\section{Discussion}

In this article, we have outlined an application of the advancements in natural language processing to a real-world problem that arises in automated scoring. Decreasing the number of fragments that are required to be reviewed is not about decreasing costs, it is about significant reductions in the amount of time required to review serious and critical threats to the safety of students. Keeping the percentage low, while maintaining a certain efficacy, and focusing on efficient models is a way of ensuring that the appropriate authorities are made aware of these life-threatening situations in a timely manner.

\bibliographystyle{plain}
\bibliography{lib}{}

\end{document}